\documentclass[letterpaper]{article}
\usepackage[preprint]{aaai2027}
\nocopyright
\usepackage[hyphens]{url}
\usepackage{graphicx}
\usepackage{natbib}
\usepackage{caption}
\IfFileExists{qtm-ts1.map}{\pdfmapfile{+qtm-ts1.map}}{}

\usepackage{booktabs}
\usepackage{amsmath}
\usepackage{amsfonts}
\usepackage{amssymb}
\usepackage{array}
\usepackage{multirow}
\usepackage{tabularx}
\usepackage{xcolor}
\usepackage{tikz}
\usetikzlibrary{arrows.meta,positioning,fit,shapes.geometric}

\title{MissDiag: Diagnostic Evaluation of Incomplete-Knowledge Robustness\\ in KGQA and KG-RAG}

\author{
Hang Wang,
Hang Dong,
Lu Liu, Chuanru Ren
}
\affiliations{}
\begin{document}
\maketitle

\begin{abstract}
Knowledge graph question answering (KGQA) and knowledge-graph-based retrieval-augmented generation (KG-RAG) aim to ground answers in explicit graph evidence, but real-world knowledge graphs are often sparse, outdated, and incomplete. Existing robustness evaluations usually report aggregate changes in answer quality after evidence is removed or perturbed, which measures sensitivity to incomplete support but leaves the source of degradation under-specified: the same score change can conflate the type of missing evidence, the response of the evaluated system, and the sensitivity of the answer-matching protocol. To address this gap, we propose \textbf{MissDiag}, a diagnostic evaluation framework for incomplete-knowledge robustness in KGQA and KG-RAG. MissDiag keeps the question and gold answer fixed while applying structurally typed missingness interventions to benchmark-provided support graphs, enabling paired comparisons that decompose robustness changes by evidence type, system response, and evaluation protocol rather than reducing them to a single aggregate score drop. Experiments across multiple system families show that incomplete-knowledge robustness is better understood as a typed degradation phenomenon than as a uniform property: answer-adjacent evidence loss produces the largest observed degradation, source-context removal is often neutral and can be beneficial, and semantic answer matching changes absolute scores while preserving the main typed degradation patterns. By transforming aggregate robustness measurement into typed diagnostic attribution, MissDiag provides a more interpretable basis for comparing, diagnosing, and stress-testing KGQA and KG-RAG systems under incomplete knowledge.

\end{abstract}

\begin{quote}
\small\textit{Code will be released after the review process.}
\end{quote}

\section{Introduction}

Knowledge graph question answering (KGQA) and knowledge-graph-based retrieval-augmented generation (KG-RAG) aim to ground answers in explicit graph evidence rather than relying only on parametric memory. This makes them important testbeds for evaluating structured reasoning, factual grounding, and answer trustworthiness. Prior work has advanced this goal through semantic parsing and query-graph construction \citep{berant2013semantic,yih2015semantic}, embedding-based and graph-based QA \citep{huang2019knowledge,sun2019pullnet,shi2021transfernet,ye-etal-2022-rng,gu-su-2022-arcaneqa,mavromatis-karypis-2022-rearev}, retrieve-and-reason systems over knowledge bases and text \citep{guu2020retrieval,lewis2020retrieval}, LLM-based KBQA \citep{xiong2024interactive}, and increasingly challenging benchmarks \citep{dubey2019lc,gu2021beyond,cao2022kqa,zhangdiagnosing}. More recently, graph-grounded LLM studies have examined whether KG augmentation improves factuality, reasoning quality, retrieval efficiency, and trustworthiness in open-ended generation settings \citep{wang2024knowledge,sun2024think,sui2025can,zhou2026breaks}. Across these lines of work, however, a persistent difficulty remains: real-world knowledge graphs are rarely complete. They are often sparse, outdated, and unevenly populated across entities and relations, making robustness under incomplete knowledge a central requirement for reliable KGQA and KG-RAG systems.

\begin{figure}[t]
\centering
\includegraphics[width=\columnwidth]{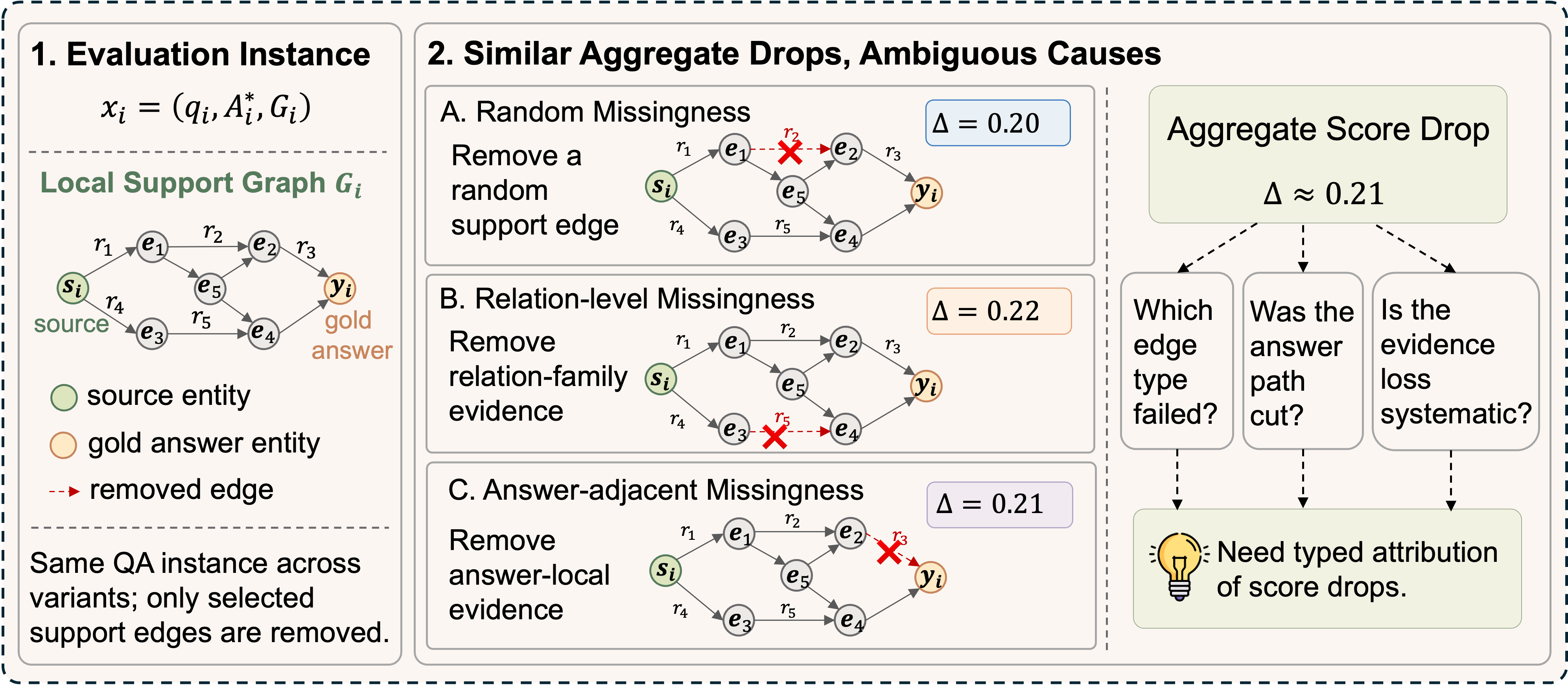}
\caption{
Motivation for typed degradation analysis. The same QA instance can show similar aggregate degradation under structurally different missingness conditions. Aggregate score drops alone do not reveal which evidence type was removed, whether answer-local evidence was affected, or whether the effect is systematic.
}
\label{fig:intro-example}
\end{figure}

Existing research has studied incomplete knowledge from several complementary perspectives, including contextual reasoning over partial graph structures \citep{mai2019contextual}, integration of textual evidence \citep{xiong2019improving,han-etal-2020-open,sun2023multi}, KG embeddings and relation prediction \citep{trouillon2016complex,schlichtkrull2018modeling,sun2019rotate,huang2019knowledge,saxena-etal-2020-improving,zhao2022improving,zan2022complex,saxena-etal-2022-sequence,guo2023knowledge}, knowledge graph completion \citep{zhao2022improving,guo2023knowledge}, completion-aware reasoning pipelines \citep{liu2022joint,ye2024rearev,han2025scr}, and LLM-based or multi-agent reasoning over incomplete graphs \citep{xu-etal-2024-generate,liu2026debate}. In parallel, evaluation studies have asked whether completion methods improve downstream QA and whether current benchmarks and metrics support reliable conclusions under incomplete or imperfect knowledge \citep{yu-etal-2023-compleqa,perevalov-etal-2022-knowledge,steinmetz2021kgqa,zhangdiagnosing,zhang2026diagnosing,zhou2026breaks,sui2025can}. These studies have substantially improved our understanding of how systems recover missing facts, exploit auxiliary evidence, or reason over partial structures. Nevertheless, current evaluation practice still has a basic attribution limitation: most evaluations remove, corrupt, or recover evidence and then report the resulting change in answer quality. Such degradation-based evaluation can measure whether performance changes, but it does not explain why the change occurs.

This ambiguity is illustrated in Figure~\ref{fig:intro-example}: the same question--answer instance may exhibit similar aggregate score drops under structurally different evidence-removal conditions, even though the underlying failure mechanisms are different. A lower score may indicate that essential supporting evidence has become unavailable; it may also indicate that the system fails to exploit retained evidence, that graph conversion or candidate construction changes the effective search space, or that the evaluation protocol fails to recognize a semantically acceptable answer. Conversely, apparent robustness or even improvement under incomplete evidence may reflect support pruning or metric behavior rather than stronger reasoning. This matters because robustness under incomplete knowledge is often used as evidence of reasoning capability and system reliability. If similar score changes can arise from different causes, then aggregate degradation alone is insufficient for interpreting robustness claims. What is needed is not only a performance measurement protocol, but a diagnostic evaluation framework that can attribute degradation to different forms of missing evidence, compare how system families respond to the same intervention, and expose how much the conclusion depends on the evaluation metric.

\begin{figure*}[t]
\centering
\includegraphics[width=\textwidth]{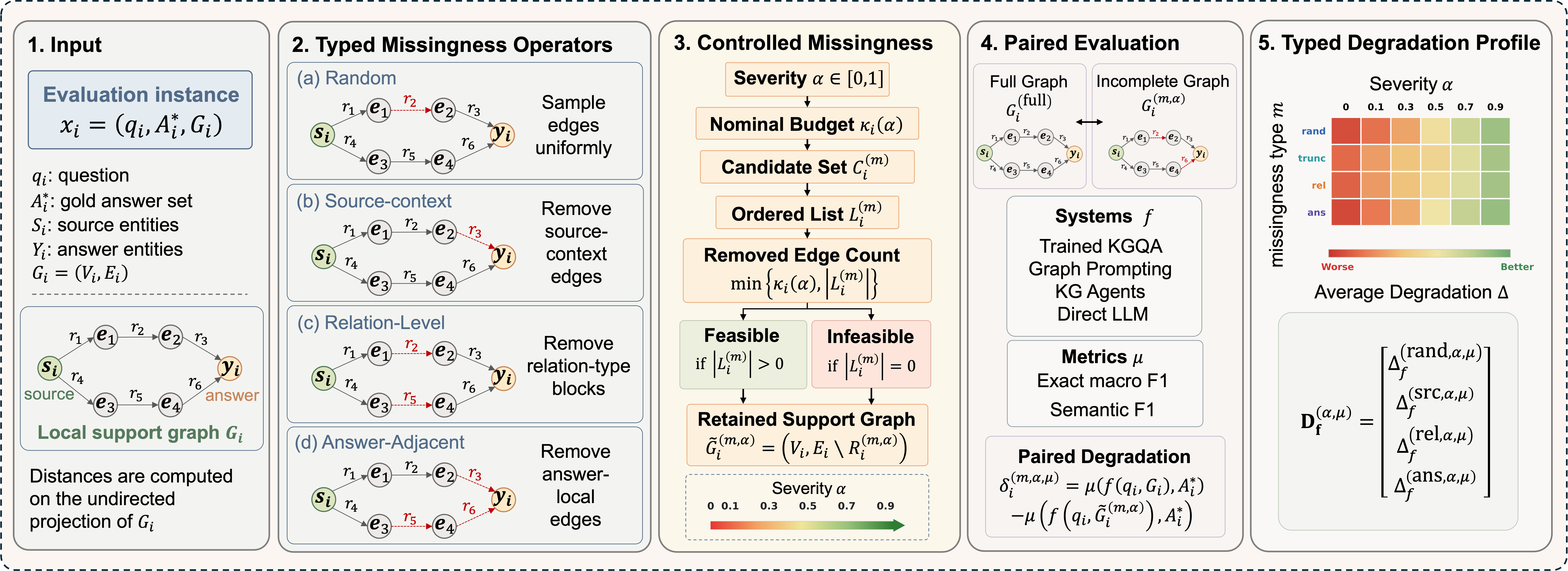}
\caption{
Overview of MissDiag. (1) \textbf{Input}: an instance contains a question, gold answer set, and local support graph. (2) \textbf{Typed Missingness Operators}: typed operators select removable support edges for random, source-context, relation-level, and answer-adjacent missingness. (3) \textbf{Severity-Controlled Missingness}: a shared severity budget produces an incomplete support graph. (4) \textbf{Paired Evaluation}: the same system is evaluated under complete and incomplete support to compute paired degradation. (5) \textbf{Typed Degradation Profile}: degradation values are summarized across missingness type, severity, system, and metric.
}
\label{fig:framework}
\end{figure*}

To address this gap, we propose \textbf{MissDiag}, a diagnostic evaluation framework for incomplete-knowledge robustness in KGQA and KG-RAG. As shown in Figure~\ref{fig:framework}, MissDiag starts from a benchmark-provided evaluation instance consisting of a question, a gold answer set, and a local support graph. It keeps the question and gold answer fixed while transforming the support graph into structurally distinct incomplete-evidence conditions. The primary missingness operators include random support loss, source-context loss, relation-level removal, and answer-adjacent removal. By evaluating the same system on the same question--answer pair before and after each typed intervention, MissDiag separates three factors that are usually entangled in incomplete-knowledge evaluation: missingness type, system response, and evaluation sensitivity.

Rather than treating incomplete knowledge as a single robustness condition, MissDiag represents it as a typed degradation phenomenon. The framework reports robustness through degradation profiles indexed by missingness type, severity, system, and evaluation metric, with structural slices used for further analysis. This design enables paired and interpretable comparison across trained KGQA models, graph-structured prompting methods, iterative KG agents, and direct LLM baselines. It therefore allows robustness claims to be examined in terms of where degradation comes from, when it reflects genuine evidence loss, and when it should instead be attributed to system behavior or evaluation sensitivity.

Our contributions are as follows:
\begin{itemize}
\item We formulate incomplete-knowledge robustness evaluation as a diagnostic attribution problem, showing that aggregate score changes conflate evidence availability, system behavior, and evaluation protocol, and therefore cannot by themselves support reliable conclusions about reasoning robustness.

\item We introduce MissDiag, a controlled diagnostic framework that applies structurally typed missingness interventions to benchmark support graphs while preserving paired question--answer comparisons, enabling degradation to be analyzed by missingness source rather than only by overall performance loss.

\item We instantiate MissDiag across multiple system families, severity levels, structural slices, and answer-matching metrics, demonstrating that the same missing-evidence intervention can lead to degradation, near invariance, or improvement depending on system design and evaluation protocol. This reveals robustness patterns that aggregate scores obscure.


\end{itemize}

\section{Related Work}

\subsection{KGQA and Graph-Grounded QA}

KGQA aims to answer natural-language questions using structured graph evidence. Early work often relied on semantic parsing or query-graph construction to map questions into executable logical forms or graph queries \citep{berant2013semantic,yih2015semantic}. Later graph-based models retrieve and reason over local evidence subgraphs, making the support structure itself part of the answering process \citep{sun2019pullnet,he2021improving,mavromatis-karypis-2022-rearev}. In parallel, retrieval-centered QA and graph-grounded LLM methods use retrieved knowledge to support generation and reasoning beyond purely parametric memory \citep{guu2020retrieval,lewis2020retrieval,wang2024knowledge,xiong2024interactive,sun2024think}. Benchmarks such as LC-QuAD~2.0, GrailQA, KQA Pro, and KGQAGen broaden evaluation beyond simple fact lookup by emphasizing compositionality, generalization, and dataset reliability \citep{dubey2019lc,gu2021beyond,cao2022kqa,zhangdiagnosing}. These studies provide the system and benchmark context for our work. MissDiag differs by treating existing systems as diagnostic subjects under controlled evidence manipulation rather than proposing another KGQA architecture.

\subsection{Incomplete-Knowledge Question Answering}

Incomplete knowledge is a persistent challenge for KGQA because real-world graphs are sparse, unevenly populated, and often missing facts needed for multi-hop reasoning. Prior work has addressed this issue by reasoning over partial graph structures \citep{mai2019contextual}, incorporating auxiliary textual evidence \citep{xiong2019improving,han-etal-2020-open}, using graph completion or completion-aware QA pipelines \citep{yu-etal-2023-compleqa,ye2024rearev}, and applying LLM-centered reasoning to incomplete graph evidence \citep{xu-etal-2024-generate,zhou2026breaks}. These approaches mainly ask how to recover or maintain answer quality when knowledge is missing. Our work asks a complementary evaluation question: when answer quality changes under incomplete knowledge, how should the change be attributed? Instead of treating missingness as a single condition, MissDiag separates different structural forms of evidence loss and compares their effects under a paired protocol.

\subsection{Benchmark Reliability and Evaluation Sensitivity}

Evaluation conclusions in KGQA and graph-grounded QA are sensitive to dataset construction, evidence availability, and answer-matching protocols. Dataset audits and leaderboard analyses show that KGQA benchmarks can contain annotation issues, heterogeneous difficulty, and inconsistent reporting practices \citep{steinmetz2021kgqa,perevalov-etal-2022-knowledge,zhangdiagnosing}. More broadly, adversarial evaluation and behavioral testing show that aggregate metrics can hide distinct failure mechanisms behind similar score changes \citep{jia2017adversarial,ribeiro2020beyond,gardner2020evaluating}. Recent KG-RAG and trustworthiness studies further suggest that graph augmentation and incomplete evidence require careful evaluation design \citep{sui2025can,zhang2026diagnosing,zhou2026breaks}. This literature motivates our view that evaluation is not a neutral reporting layer. MissDiag extends this line of work by decomposing incomplete-knowledge evaluation into missingness type, system response, severity, and metric sensitivity, so that robustness claims can be interpreted beyond a single aggregate degradation score.
\section{Method}
\label{sec:method}

This section introduces MissDiag, a diagnostic evaluation framework for incomplete-knowledge robustness in KGQA and KG-RAG. As shown in Figure~\ref{fig:framework}, the framework keeps the question, gold answer, and evaluated system fixed, transforms the support graph with severity-controlled typed missingness operators, and compares the resulting outputs against the complete-support condition. This design turns aggregate robustness changes into paired degradation profiles indexed by missingness type, severity, system, and metric. The section first formulates the diagnostic evaluation problem, then describes support graph construction, defines the missingness operators, and presents the paired degradation profile used as the main diagnostic output.

\subsection{Diagnostic Formulation}

Each evaluation instance is represented as
\begin{equation}
x_i = (q_i, A_i^\ast, G_i),
\label{eq:instance}
\end{equation}
where \(q_i\) is the question, \(A_i^\ast\) is the gold answer set, and \(G_i=(V_i,E_i)\) is the local support graph. The source entities linked to the question are denoted by \(S_i \subseteq V_i\), and the gold-answer entities aligned to graph nodes are denoted by \(Y_i \subseteq V_i\). Operators requiring unavailable alignments mark the instance infeasible for the corresponding operator. Given an evaluated system \(f\), MissDiag compares the complete-support prediction \(f(q_i,G_i)\) with predictions obtained after transforming only the support edges. Across conditions, the question, gold answer, source entities, and node inventory remain fixed; only the retained edge evidence changes.

\subsection{Support Graph Construction}

The local support graph \(G_i\) is constructed from the evidence associated with instance \(x_i\). When the evidence is given as triples, proof paths, or a retrieved subgraph, it is converted into a labeled directed graph. Graph distances used by the missingness operators are computed on the undirected projection of this graph. For a node \(u\) and node set \(B\), \(\mathrm{dist}(u,B)\) denotes the shortest-path distance from \(u\) to any node in \(B\) on the undirected projection.

The edge set is written as
\begin{equation}
E_i = E_i^{\mathrm{sup}} \cup E_i^{\mathrm{ctx}},
\label{eq:support-edges}
\end{equation}
where \(E_i^{\mathrm{sup}}\) denotes the original support evidence and \(E_i^{\mathrm{ctx}}\) denotes a possibly empty set of optional source-anchored context edges. The context edges are constructed before any missingness intervention, follow a deterministic selection rule, and remain fixed across all conditions. This ensures that all missingness operators start from the same support graph.

After an intervention, the evaluated system receives only the retained support evidence. The node inventory may be kept fixed internally for alignment and paired comparison, but isolated nodes are not exposed as additional answer hints in generative KG-RAG prompts. Additional construction details are provided in the supplementary material.

\subsection{Severity-Controlled Missingness}

Let \(\alpha \in [0,1]\) denote the missingness severity and \(n_i=|E_i|\) the number of support edges. For each instance, the nominal removal budget is defined as
\begin{equation}
\kappa_i(\alpha)
=
\begin{cases}
0, & n_i \le 1 \text{ or } \alpha = 0,\\
\min\bigl(n_i-1,\, \eta_i(\alpha)\bigr), & \text{otherwise,}
\end{cases}
\label{eq:budget}
\end{equation}
where \(\eta_i(\alpha)=\max(1,\mathrm{round}(\alpha n_i))\). This budget preserves complete support at \(\alpha=0\) and keeps at least one support edge when removal is applied.

Each missingness operator \(m\) defines an ordered candidate edge list \(L_i^{(m)}\). Given the shared budget, the removed edge set \(R_i^{(m,\alpha)}\) consists of the first \(\min(\kappa_i(\alpha), |L_i^{(m)}|)\) edges in \(L_i^{(m)}\). The retained support graph is
\begin{equation}
\widetilde{G}_i^{(m,\alpha)}
=
(V_i, E_i \setminus R_i^{(m,\alpha)}),
\label{eq:retained-graph}
\end{equation}
where \(R_i^{(m,\alpha)}\) is the removed edge set. If \(\kappa_i(\alpha)>0\) but \(L_i^{(m)}\) is empty, the instance is marked infeasible for operator \(m\).

\subsection{Typed Missingness Operators}

MissDiag uses four typed missingness operators: random, source-context, relation-level, and answer-adjacent missingness. Each operator first defines a candidate edge set and then orders it into \(L_i^{(m)}\), which is used by the shared removal budget. The operators are designed to probe different structural forms of evidence loss under the same removal budget, rather than to define mutually exclusive edge categories.

For an edge \(e=(u,r,v)\), two structural scores are used:

\begin{equation}
\tau_i(e)
=
\max\{\mathrm{dist}(u,S_i),\mathrm{dist}(v,S_i)\},
\label{eq:source-depth}
\end{equation}
\begin{equation}
\rho_i(e)
=
\min\{\mathrm{dist}(u,Y_i),\mathrm{dist}(v,Y_i)\}.
\label{eq:answer-distance}
\end{equation}
Here, \(\tau_i(e)\) measures source depth and \(\rho_i(e)\) measures answer distance.

\paragraph{Random missingness.}
Random missingness uses all support edges as candidates, \(C_i^{(\mathrm{rand})}=E_i\). The ordered list \(L_i^{(\mathrm{rand})}\) is a uniform random permutation of \(C_i^{(\mathrm{rand})}\). This operator serves as a quantity-matched baseline for generic support loss.

\paragraph{Source-context missingness.}
Source-context missingness targets edges incident to source entities but not incident to aligned answer entities:
\begin{equation}
\begin{aligned}
C_i^{(\mathrm{src})}
=
\{e=(u,r,v)\in E_i:\;&
\{u,v\}\cap S_i\neq\emptyset,\\
&
\{u,v\}\cap Y_i=\emptyset\}.
\end{aligned}
\label{eq:source-candidates}
\end{equation}
The ordered list \(L_i^{(\mathrm{src})}\) is obtained by sorting candidates by increasing \(\tau_i(e)\), with deterministic tie-breaking. This operator probes sensitivity to source-side context while excluding answer-incident evidence from its candidate set.

\paragraph{Relation-level missingness.}
Relation-level missingness uses all support edges as candidates, \(C_i^{(\mathrm{rel})}=E_i\), and groups them by relation type. Relation blocks are ordered by decreasing frequency in \(E_i\), and edges are removed block by block under the shared budget. This operator probes sensitivity to relation-family evidence.

\paragraph{Answer-adjacent missingness.}
Answer-adjacent missingness targets edges incident to, or one hop from, aligned answer entities:
\begin{equation}
C_i^{(\mathrm{ans})}
=
\{e\in E_i:\rho_i(e)\le 1\}.
\label{eq:answer-candidates}
\end{equation}
The ordered list \(L_i^{(\mathrm{ans})}\) is obtained by sorting candidates by increasing \(\rho_i(e)\), with deterministic tie-breaking. This operator probes sensitivity to answer-local grounding evidence.

\subsection{Paired Degradation Profiles}

Let \(\mu\) be the answer-set evaluation metric used for a given comparison. MissDiag treats the metric as an explicit axis of the diagnostic profile. For system \(f\), the complete-support score of instance \(i\) is \(\mu(f(q_i,G_i), A_i^\ast)\), and the score under missingness type \(m\) and severity \(\alpha\) is \(\mu(f(q_i,\widetilde{G}_i^{(m,\alpha)}), A_i^\ast)\). The paired degradation is defined as
\begin{equation}
\delta_i^{(m,\alpha,\mu)}
=
\mu(f(q_i,G_i), A_i^\ast)
-
\mu(f(q_i,\widetilde{G}_i^{(m,\alpha)}), A_i^\ast).
\label{eq:paired-degradation}
\end{equation}
Positive values indicate performance loss after evidence removal, while negative values indicate improvement.

Let \(\mathcal{I}^{(m,\alpha)}\) be the feasible instance set for operator \(m\) at severity \(\alpha\). The dataset-level degradation of system \(f\) is
\begin{equation}
\Delta_f^{(m,\alpha,\mu)}
=
\frac{1}{|\mathcal{I}^{(m,\alpha)}|}
\sum_{i \in \mathcal{I}^{(m,\alpha)}}
\delta_i^{(m,\alpha,\mu)}.
\label{eq:dataset-degradation}
\end{equation}

The main diagnostic output is the typed degradation profile
\begin{equation}
\mathbf{D}_f^{(\alpha,\mu)}
=
\begin{bmatrix}
\Delta_f^{(\mathrm{rand},\alpha,\mu)}\\
\Delta_f^{(\mathrm{src},\alpha,\mu)}\\
\Delta_f^{(\mathrm{rel},\alpha,\mu)}\\
\Delta_f^{(\mathrm{ans},\alpha,\mu)}
\end{bmatrix}.
\label{eq:profile}
\end{equation}
This profile summarizes how a fixed system degrades under different missingness types at a given severity and metric. Instances infeasible for an operator are excluded from that operator's aggregation, with feasibility details provided in the supplementary material.

\section{Experiments}
\label{sec:experiments}

We evaluate whether MissDiag provides a more informative view of incomplete-knowledge robustness than a single aggregate score. Following the method design, the evaluation first compares typed degradation profiles across system families, then examines severity effects, metric sensitivity, and structural slices. The experiments are organized around four research questions:

\begin{itemize}
    \item \textbf{RQ1:} Do missingness types produce distinct degradation profiles?
    \item \textbf{RQ2:} How do profiles change as severity increases?
    \item \textbf{RQ3:} Do profiles depend on the evaluation metric?
    \item \textbf{RQ4:} When is answer-adjacent degradation strongest?
\end{itemize}

\subsection{Experimental Setup}
\label{sec:exp-setup}

\paragraph{Data.}
We use KGQAGen-10k~\citep{zhangdiagnosing} as the main evaluation benchmark. Its instances provide question--answer pairs and local evidence structures, which are converted into the support-graph format used by MissDiag. We link source entities from the question and align gold-answer entities to graph nodes when available. The main experiments use the development split. Dataset-specific support construction details are provided in the supplementary material.

\paragraph{Evaluation protocol.}
We use a paired fixed-input protocol. For each instance, the question, gold answer set, source entities, and node inventory are kept fixed across complete and incomplete-support conditions; only the retained support edges change. For a fixed instance, missingness type, and severity, all systems are evaluated on the same transformed support graph. The main cross-system comparison uses 1,050 examples for which the complete-support condition and all four missingness conditions are feasible. LLM-focused analyses use the corresponding feasible subset, with sample sizes reported in the relevant tables.

\paragraph{Systems.}
We evaluate four system families:
(1) \textit{Trained KGQA models}, including ReaRev~\citep{mavromatis-karypis-2022-rearev}, NuTrea~\citep{choi2023nutrea}, and NSM~\citep{he2021improving};
(2) \textit{Graph prompting methods}, including MindMap~\citep{wen-etal-2024-mindmap}, StructGPT~\citep{jiang-etal-2023-structgpt}, and KG-GPT~\citep{kim-etal-2023-kg};
(3) \textit{KG agents}, including ToG~\citep{sun2024think}, PoG~\citep{chen2024pog}, and GoG~\citep{xu-etal-2024-generate}; and
(4) \textit{Direct LLM baselines}, including Qwen2.5-7B-Instruct\footnote{\url{https://huggingface.co/Qwen/Qwen2.5-7B-Instruct}}, Qwen2.5-14B-Instruct\footnote{\url{https://huggingface.co/Qwen/Qwen2.5-14B-Instruct}}, Llama-3.1-8B-Instruct\footnote{\url{https://huggingface.co/meta-llama/Llama-3.1-8B-Instruct}}, and Mistral-7B-Instruct-v0.3\footnote{\url{https://huggingface.co/mistralai/Mistral-7B-Instruct-v0.3}}.
For graph prompting methods and KG agents, all methods use Qwen2.5-7B-Instruct as the shared LLM backbone across complete-support and missingness conditions. This controls for backbone capability and focuses the comparison on prompting, planning, and agent workflow. For all LLM-based systems, graph access is restricted to the transformed local support graph.

\paragraph{Missingness conditions.}
We compare complete support with four typed missingness conditions: random, source-context, relation-level, and answer-adjacent missingness. Unless otherwise stated, the main comparison uses severity \(\alpha=0.3\). The severity analysis evaluates \(\alpha \in \{0.1,0.3,0.5\}\).

\paragraph{Metrics and reporting.}
The primary metric is exact macro set-level F1. Results are reported as complete-support F1 and paired degradation \(\Delta\)F1, where positive values indicate performance loss after evidence removal and negative values indicate improvement. Metric sensitivity is evaluated by comparing exact F1 with semantic F1 for direct LLM baselines. Semantic F1 replaces exact string matching with semantic equivalence matching before computing set-level precision and recall. Additional low-level details, including prompt templates and feasibility bookkeeping, are provided in the supplementary material.

\begin{table*}[t]
\centering
\small
\setlength{\tabcolsep}{6pt}
\renewcommand{\arraystretch}{1.12}
\begin{tabular*}{\textwidth}{
@{\hspace{0.8em}}l
@{\hspace{1.8em}}
l
@{\extracolsep{\fill}}
ccccc
@{\hspace{0.8em}}
}
\toprule
\multirow{2}{*}{\textbf{Category}}
& \multirow{2}{*}{\textbf{System}}
& \multicolumn{1}{c}{\textbf{Complete}}
& \multicolumn{4}{c}{\textbf{Paired degradation \(\boldsymbol{\Delta}\)F1}} \\
\cmidrule(lr){3-3}\cmidrule(lr){4-7}
& & \textbf{F1}
& \textbf{Random}
& \textbf{Source-context}
& \textbf{Relation-level}
& \textbf{Answer-adjacent} \\
\midrule
\multirow{3}{*}{\shortstack[l]{\textbf{Trained KGQA}\\\textbf{Models}}}
& ReaRev
& 80.8 & 4.5 & -1.2 & 2.7 & 11.9 \\
& NuTrea
& 77.2 & 1.9 & -9.9 & -2.7 & 12.2 \\
& NSM
& 46.7 & 3.0 & -24.1 & -5.2 & 12.1 \\
\midrule
\multirow{3}{*}{\shortstack[l]{\textbf{Graph Prompting}\\\textbf{Methods}}}
& MindMap
& 46.2 & 10.1 & 0.6 & 11.2 & 12.7 \\
& StructGPT
& 52.9 & 9.6 & -0.7 & 8.0 & 13.8 \\
& KG-GPT
& 45.7 & 7.0 & 4.2 & 5.3 & 10.3 \\
\midrule
\multirow{3}{*}{\textbf{KG Agents}}
& ToG
& 79.5 & 7.0 & 3.9 & 6.0 & 11.4 \\
& PoG
& 79.9 & 6.8 & 2.7 & 5.8 & 11.6 \\
& GoG
& 81.7 & 12.1 & 2.3 & 8.9 & 19.8 \\
\midrule
\multirow{4}{*}{\shortstack[l]{\textbf{Direct LLM}\\\textbf{Baselines}}}
& Qwen2.5-7B-Instruct
& 85.3 & 11.1 & 3.0 & 7.4 & 21.3 \\
& Qwen2.5-14B-Instruct
& 90.0 & 6.7 & 1.2 & 5.5 & 17.0 \\
& Llama-3.1-8B-Instruct
& 84.4 & 6.1 & -0.5 & 4.0 & 15.8 \\
& Mistral-7B-Instruct-v0.3
& 81.2 & 5.4 & 1.6 & 3.8 & 13.3 \\
\bottomrule
\end{tabular*}
\caption{
Main typed degradation profiles across system families at \(\alpha=0.3\). Scores are reported as complete-support macro F1 and paired \(\Delta\)F1 under each missingness type. Positive \(\Delta\)F1 indicates degradation; negative values indicate improvement.
}
\label{tab:main-profile}
\end{table*}

\subsection{Implementation Details}
\label{sec:implementation-details}

Random missingness uses a fixed global seed, with per-example seeds derived from the sample identifier, severity, and operator. Non-random operators use deterministic tie-breaking when ordering candidate edges. ReaRev, NuTrea, and NSM are trained once on complete-support data using pinned official implementations, and their best-F1 checkpoints are reused across all missingness conditions without condition-specific retraining. For LLM-based systems, retained graph edges are serialized as textual triples and provided as the only graph evidence in the prompt. Inference uses deterministic greedy decoding without sampling, with fixed generation and reasoning budgets across evidence conditions. Experiments are conducted using NVIDIA GH200 GPUs; LLM-based experiments use PyTorch 2.9.1, CUDA 12.8, Transformers 4.46.2, and bfloat16 inference. Additional implementation details are provided in the supplementary material.
\subsection{RQ1: Typed Degradation Profiles Across Systems}
\label{sec:rq1-profiles}

Table~\ref{tab:main-profile} reports the main cross-system comparison on 1,050 paired development examples at \(\alpha=0.3\). Each row gives the complete-support F1 of a system and its paired \(\Delta\)F1 under the four missingness conditions, directly instantiating the typed degradation profile defined in the method.
Three patterns are clear: First, answer-adjacent removal is the dominant degradation condition for every system, with drops ranging from 10.3 to 21.3 F1 points. This shows that answer-local evidence loss is consistently more damaging than generic support removal across trained KGQA models, graph prompting methods, KG agents, and direct LLM baselines. Second, source-context removal behaves differently: it is small for several prompting and direct LLM systems, and negative for several trained KGQA models. This indicates that removing source-side context can sometimes reduce distracting evidence rather than harm prediction. Third, random and relation-level removal usually produce intermediate degradation, suggesting that support quantity and relation-family loss affect performance but do not explain the full degradation pattern.

Together, these results provide the main empirical support for typed degradation profiles: under the same paired protocol, incomplete-support degradation depends on what kind of evidence is missing and how each system uses the remaining graph. A single aggregate degradation score would collapse these distinct effects and obscure the difference between harmful answer-local loss, neutral or beneficial source-context removal, and intermediate random or relation-level loss.

\subsection{RQ2: Severity Effects}
\label{sec:rq2-severity}

\begin{figure}[t]
\centering
\IfFileExists{figures/severity_curves.pdf}{%
\includegraphics[width=\columnwidth]{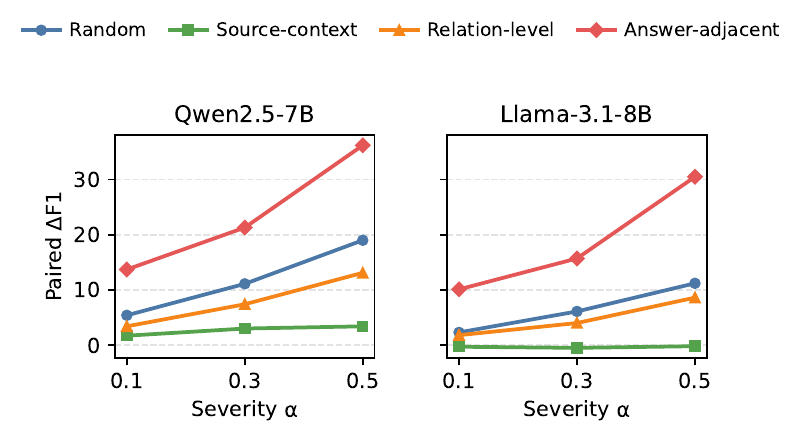}%
}{%
\fbox{\parbox[c][1.75in][c]{0.95\columnwidth}{\centering Severity-curves placeholder\\[2pt]
\small Missing file: figures/severity\_curves.pdf}}%
}
\caption{
Severity effects on paired degradation. Curves show paired \(\Delta\)F1 across missingness types for Qwen2.5-7B-Instruct and Llama-3.1-8B-Instruct as \(\alpha\) increases.
}
\label{fig:severity-curves}
\end{figure}

Figure~\ref{fig:severity-curves} reports severity-dependent degradation for representative direct LLM baselines. The severity parameter varies over \(\alpha \in \{0.1,0.3,0.5\}\), while the question, gold answer, metric, and missingness operators remain fixed.
The curves show that increasing severity amplifies degradation, but not uniformly across missingness types. Answer-adjacent removal remains the strongest degradation condition at every severity and grows most sharply as \(\alpha\) increases. Random and relation-level removal also increase with severity, but their degradation remains consistently below answer-adjacent removal. Source-context removal stays comparatively small and changes little across severity levels.
These results show that the typed degradation pattern is not an artifact of the main \(\alpha=0.3\) setting. Severity controls the scale of evidence loss, while the relative behavior of missingness types remains structurally distinct.

\subsection{RQ3: Metric Sensitivity}
\label{sec:rq3-metric}

Table~\ref{tab:metric-sensitivity} compares paired degradation under exact F1 and semantic F1 for direct LLM baselines at \(\alpha=0.3\). This analysis tests whether the typed degradation profile changes when answer matching allows semantic equivalence rather than exact surface matching.
The results show that metric choice changes degradation magnitudes but not the main typed pattern. For all four direct LLM baselines, answer-adjacent removal remains the largest degradation condition under both exact F1 and semantic F1. Semantic matching slightly changes individual \(\Delta\)F1 values, but it does not reverse the ordering of missingness effects. This indicates that the main diagnostic conclusion depends primarily on the type of missing evidence rather than on the particular answer-matching rule.

\begin{table}[t]
\centering
\small
\setlength{\tabcolsep}{3.5pt}
\renewcommand{\arraystretch}{1.12}
\begin{tabular*}{\columnwidth}{@{\hspace{0.3em}}l@{\extracolsep{\fill}}lrrrr@{\hspace{0.3em}}}
\toprule
\multirow{2}{*}{\textbf{System}}
& \multirow{2}{*}{\textbf{Metric}}
& \multicolumn{4}{c}{\textbf{Paired degradation \(\boldsymbol{\Delta}\)F1}} \\
\cmidrule(lr){3-6}
& & \textbf{Rand} & \textbf{Src} & \textbf{Rel} & \textbf{Ans-adj} \\
\midrule
\multirow{2}{*}{Qwen2.5-7B}
& Exact F1    & 11.1 & 3.0 & 7.4 & 21.3 \\
& Semantic F1 & 11.6 & 3.2 & 8.0 & 21.4 \\
\midrule
\multirow{2}{*}{Qwen2.5-14B}
& Exact F1    & 6.7 & 1.2 & 5.5 & 17.0 \\
& Semantic F1 & 6.9 & 1.2 & 5.4 & 16.1 \\
\midrule
\multirow{2}{*}{Llama-3.1-8B}
& Exact F1    & 6.1 & -0.5 & 4.0 & 15.8 \\
& Semantic F1 & 5.1 & -1.0 & 3.9 & 15.1 \\
\midrule
\multirow{2}{*}{Mistral-7B}
& Exact F1    & 5.4 & 1.6 & 3.8 & 13.3 \\
& Semantic F1 & 5.1 & 0.9 & 3.2 & 13.5 \\
\bottomrule
\end{tabular*}
\caption{
Metric sensitivity of typed degradation profiles for direct LLM baselines at \(\alpha=0.3\). Rand, Src, Rel, and Ans-adj denote random, source-context, relation-level, and answer-adjacent missingness.
}
\label{tab:metric-sensitivity}
\end{table}

\subsection{RQ4: Structural Slices}
\label{sec:rq4-slices}

Table~\ref{tab:slice-analysis} examines when answer-adjacent degradation is strongest. The analysis compares random and answer-adjacent degradation across answer cardinality and support-graph size, averaged over the four direct LLM baselines. The gap is defined as answer-adjacent \(\Delta\)F1 minus random \(\Delta\)F1.

\begin{table}[t]
\centering
\scriptsize
\setlength{\tabcolsep}{3.8pt}
\renewcommand{\arraystretch}{1.12}
\resizebox{\columnwidth}{!}{%
\begin{tabular}{llrrrrr}
\toprule
\textbf{Slice}
& \textbf{Group}
& \textbf{N}
& \textbf{Full}
& \textbf{Rand}
& \textbf{Ans-adj}
& \textbf{Gap} \\
\midrule
\multirow{2}{*}{Answer cardinality}
& Single-answer & 888 & 90.8 & 6.8 & 14.4 & 7.6 \\
& Multi-answer  & 163 & 54.5 & 10.4 & 30.2 & 19.9 \\
\midrule
\multirow{3}{*}{Support size}
& Small support  & 353 & 80.5 & 12.6 & 26.7 & 14.2 \\
& Medium support & 527 & 87.0 & 5.2  & 13.3 & 8.0 \\
& Large support  & 171 & 89.2 & 2.9  & 7.4  & 4.5 \\
\bottomrule
\end{tabular}
}
\caption{
Structural slices for direct LLM baselines at \(\alpha=0.3\). Rand and Ans-adj report paired \(\Delta\)F1 under random and answer-adjacent missingness. Gap is the difference between Ans-adj and Rand degradation.
}
\label{tab:slice-analysis}
\end{table}

Answer-adjacent degradation is most pronounced for multi-answer questions and small-support graphs. Multi-answer examples show a 19.9-point gap over random removal, compared with 7.6 points for single-answer examples. The gap also decreases as support size grows, from 14.2 points on small-support graphs to 4.5 points on large-support graphs. These results show that answer-local evidence loss is especially harmful when answers are structurally harder to recover or when alternative support paths are limited.

\section{Discussion}

Taken together, the experiments show that incomplete-knowledge robustness is better characterized by typed degradation profiles than by a single aggregate score. The dominant effect comes from answer-adjacent evidence loss, while severity, metric, and structural analyses clarify when this effect is amplified or preserved. A system can appear robust under one missingness type while being highly sensitive to another. This is most visible in the contrast between source-context and answer-adjacent removal. Source-context removal is often small or even beneficial, suggesting that some source-side evidence may introduce distraction or increase the burden of evidence selection. In contrast, answer-adjacent removal consistently produces the largest degradation, indicating that systems depend strongly on evidence near the aligned answer entities. These two effects would be collapsed by an aggregate missing-evidence score, even though they imply different failure mechanisms.

Typed profiles also make cross-system comparison more interpretable. Higher complete-support F1 does not necessarily imply stronger robustness under all forms of missingness. For example, direct LLM baselines achieve strong complete-support performance but can still show large answer-adjacent degradation. Conversely, some trained KGQA models show negative degradation under source-context removal, indicating that their behavior under incomplete evidence is not captured by complete-support accuracy alone. This highlights MissDiag's diagnostic role in revealing how systems respond to different evidence losses, not only how much their average score changes.

\section{Conclusion}

This paper studies incomplete-knowledge KGQA and KG-RAG evaluation as a diagnostic problem. A score drop under incomplete knowledge is not directly interpretable because it can conflate missing evidence, system response, and evaluation protocol. MissDiag addresses this ambiguity by keeping the question, gold answer, source entities, and node inventory fixed while applying severity-controlled typed missingness operators to support edges. Across trained KGQA models, graph prompting methods, KG agents, and direct LLM baselines, answer-adjacent evidence loss produces the most consistent degradation, whereas source-context removal can be neutral or beneficial. Severity, metric, and structural analyses further show when this typed pattern is preserved or amplified. The central implication is methodological: incomplete-knowledge robustness should be reported as typed degradation profiles, rather than as one aggregate score drop.

\bibliography{refs}

\end{document}